\documentclass[10pt,twocolumn,letterpaper]{article}

\usepackage{cvpr}
\usepackage{times}
\usepackage{epsfig}
\usepackage{graphicx}
\usepackage{amsmath}
\usepackage{amssymb}

 \cvprfinalcopy 

\def\cvprPaperID{1} 

\ifcvprfinal\fi
\begin{document}

\title{Compressive Light Field Reconstructions using Deep Learning}

\author{Mayank Gupta\footnotemark[1]\\
Arizona State University\\
\and
Arjun Jauhari\footnotemark[1]\\
Cornell University\\
\and
Kuldeep Kulkarni\\
Arizona State University\\
\and
Suren Jayasuriya\\
Carnegie Mellon University\\
\and
Alyosha Molnar\\
Cornell University\\
\and
Pavan Turaga\\
Arizona State University\\
}

\maketitle
\footnotetext[1]{Authors contributed equally to this paper.}

\begin{abstract}
Light field imaging is limited in its computational processing demands of high sampling for both spatial and angular dimensions. Single-shot light field cameras sacrifice spatial resolution to sample angular viewpoints, typically by multiplexing incoming rays onto a 2D sensor array. While this resolution can be recovered using compressive sensing, these iterative solutions are slow in processing a light field. We present a deep learning approach using a new, two branch network architecture, consisting jointly of an autoencoder and a 4D CNN, to recover a high resolution 4D light field from a single coded 2D image. This network decreases reconstruction time significantly while achieving average PSNR values of 26-32 dB on a variety of light fields. In particular, reconstruction time is decreased from 35 minutes to 6.7 minutes as compared to the dictionary method for equivalent visual quality. These reconstructions are performed at small sampling/compression ratios as low as 8\%, allowing for cheaper coded light field cameras. We test our network reconstructions on synthetic light fields, simulated coded measurements of real light fields captured from a Lytro Illum camera, and real coded images from a custom CMOS diffractive light field camera. The combination of compressive light field capture with deep learning allows the potential for real-time light field video acquisition systems in the future.
\end{abstract}

\section{Introduction}
Light fields, 4D representations of light rays in unoccluded space, are ubiquitous in computer graphics and vision. Light fields have been used for novel view synthesis~\cite{levin2010linear}, synthesizing virtual apertures for images post-capture~\cite{levoy2006light}, and 3D depth mapping and shape estimation~\cite{tao2017shape}. Recent research has used light fields as the raw input for visual recognition algorithms such as identifying materials~\cite{wang20164d}. Finally, biomedical microscopy has employed light field techniques to improve issues concerning aperture and depth focusing~\cite{levoy2006microscopy}. 

While the algorithmic development for light fields has yielded promising results, capturing high resolution 4D light fields at video rates is difficult. For dense sampling of the angular views, bulky optical setups involving gantries, mechanical arms, or camera arrays have been introduced~\cite{wilburn2005high,venkataraman2013picam}. However, these systems either cannot operate in real-time or must process large amounts of data, preventing deployment on embedded vision platforms with tight energy budgets. In addition, small form factor, single-shot light field cameras such as pinhole or microlens arrays above image sensors sacrifice spatial resolution for angular resolution in a fixed trade-off~\cite{codedmask, microlenses}. Even the Lytro Illum, the highest resolution consumer light field camera available, does not output video at 30 fps or higher. There is a clear need for a small form-factor, low data rate, cheap light field camera that can process light field video data efficiently. 

To reduce the curse of dimensionality when sampling light fields, we turn to compressive sensing (CS). CS states that it is possible to reconstruct a signal perfectly from small number of linear measurements, provided the number of measurements is sufficiently large, and the signal is sparse in a transform domain. Thus CS provides a principled way to reduce the amount of data that is sensed and transmitted through a communication channel. Moreover, the number of sensor elements also reduces significantly, paving a way for cheaper imaging. Recently, researchers introduced \textit{compressive light field photography} to reconstruct light fields captured from coded aperture/mask based cameras at high resolution~\cite{marwah2013compressive}. The key idea was to use dictionary-based learning for local light field atoms (or patches) coupled with sparsity-constrained optimization to recover the missing information. However, this technique required extensive computational processing on the order of hours for each light field.

In this paper, we present a new class of solutions for the recovery of compressive light fields at a fraction of the time-complexity of the current state-of-the-art, while delivering comparable (and sometimes even better) PSNR. We leverage hybrid deep neural network architectures that draw inspiration from simpler architectures in 2D inverse problems, but are redesigned for 4D light fields. We propose a new network architecture consisting of a traditional autoencoder and a 4D CNN which can invert several types of compressive light field measurements including those obtained from coded masks~\cite{codedmask} and Angle Sensitive Pixels~\cite{wang2012light, asp}. We benchmark our network reconstructions on simulated light fields, simulated compressive capture from real Lytro Illum light fields provided by Kalantari \etal~\cite{LearningViewSynthesis}, and real images from a prototype ASP camera~\cite{asp}. We achieve processing times on the order of a few minutes, which is an order of magnitude faster than the dictionary-based method. This work can help bring real-time light field video at high spatial resolution closer to reality.

\section{Related Work}
\textbf{Light Fields and Capture Methods:} The modern formulation of light fields were first introduced independently by Levoy and Hanrahan~\cite{levoy1996light} and Gortler \etal~\cite{gortler1996lumigraph}. Since then, there has been numerous work in view synthesis, synthetic aperture imaging, and depth mapping, see~\cite{levoy2006light} for a broad overview. For capture, gantries or camera arrays~\cite{wilburn2005high,venkataraman2013picam} provide dense sampling while single-shot camera methods such as microlenses~\cite{microlenses}, coded apertures~\cite{levin2007image}, masks~\cite{codedmask}, diffractive pixels~\cite{asp}, and even diffusers~\cite{antipa2016single} and random refractive water droplets~\cite{wender:2015} have been proposed. All these single-shot methods multiplex angular rays into spatial bins, and thus need to recover that lost information in post-processing. 

\textbf{Light Field Reconstruction:} Several techniques have been proposed to increase the spatial and angular resolution of captured light fields. These include using explicit signal processing priors~\cite{levin2010linear} and frequency domain methods~\cite{shi2014light}. The work closest to our own is compressive light field photography~\cite{marwah2013compressive} that uses learned dictionaries to reconstruct light fields, and extending that technique to Angle Sensitive Pixels~\cite{asp}. We replace their framework by using deep learning to perform both the feature extraction and reconstruction with a neural network. Similar to our work, researchers have recently used deep learning networks for view synthesis~\cite{LearningViewSynthesis} and spatio-angular superresolution~\cite{yoon2015learning}. However, all these methods start from existing 4D light fields, and thus they do not recover light fields from compressed or multiplexed measurements. Recently, Wang \etal proposed a hybrid camera system consisting of a DSLR camera at 30 fps with a Lytro Illum at 3fps, and used deep learning to recover light field video at 30 fps~\cite{wang2017light}. Our work hopes to make light field video processing cheaper by decreasing the spatio-angular measurements needed at capture time. 

\textbf{Compressive Sensing:} There have been numerous works in compressed sensing~\cite{candes2008introduction} resulting in various algorithms to recover the original signal. The classical algorithms \cite{donoho2006compressed,candes2006near, candes2006robust} rely on the assumption that the signal is sparse or compressible in transform domains like wavelets, DCT, or data dependent pre-trained dictionaries. More sophisticated algorithms include model-based methods~\cite{baraniuk2010model, kim2010compressed} and message-passing algorithms~\cite{donoho2009message} which impose a complex image model to perform reconstruction. However, all of these algorithms are iterative and hence are not conducive for fast reconstruction. Similar to our work, deep learning has been used for recovering 2D images from compressive measurements at faster speeds than iterative solvers. Researchers have proposed stacked-denoising autoencoders to perform CS image and video reconstruction respectively~\cite{mousavi2015deep, illiadis2016deep}. In contrast, Kulkarni \etal show that CNNs, which are traditionally used for inference tasks, can also be used for CS image reconstruction~\cite{reconnet} . We marry the benefits of the two types of architectures mentioned above and propose a novel architecture to 4D light fields that introduce additional challenges and opportunities for deep learning + compressive sensing.

\section{Light Field Photography}
\begin{figure}[t]
\begin{center}
 \includegraphics[width=0.8\linewidth]{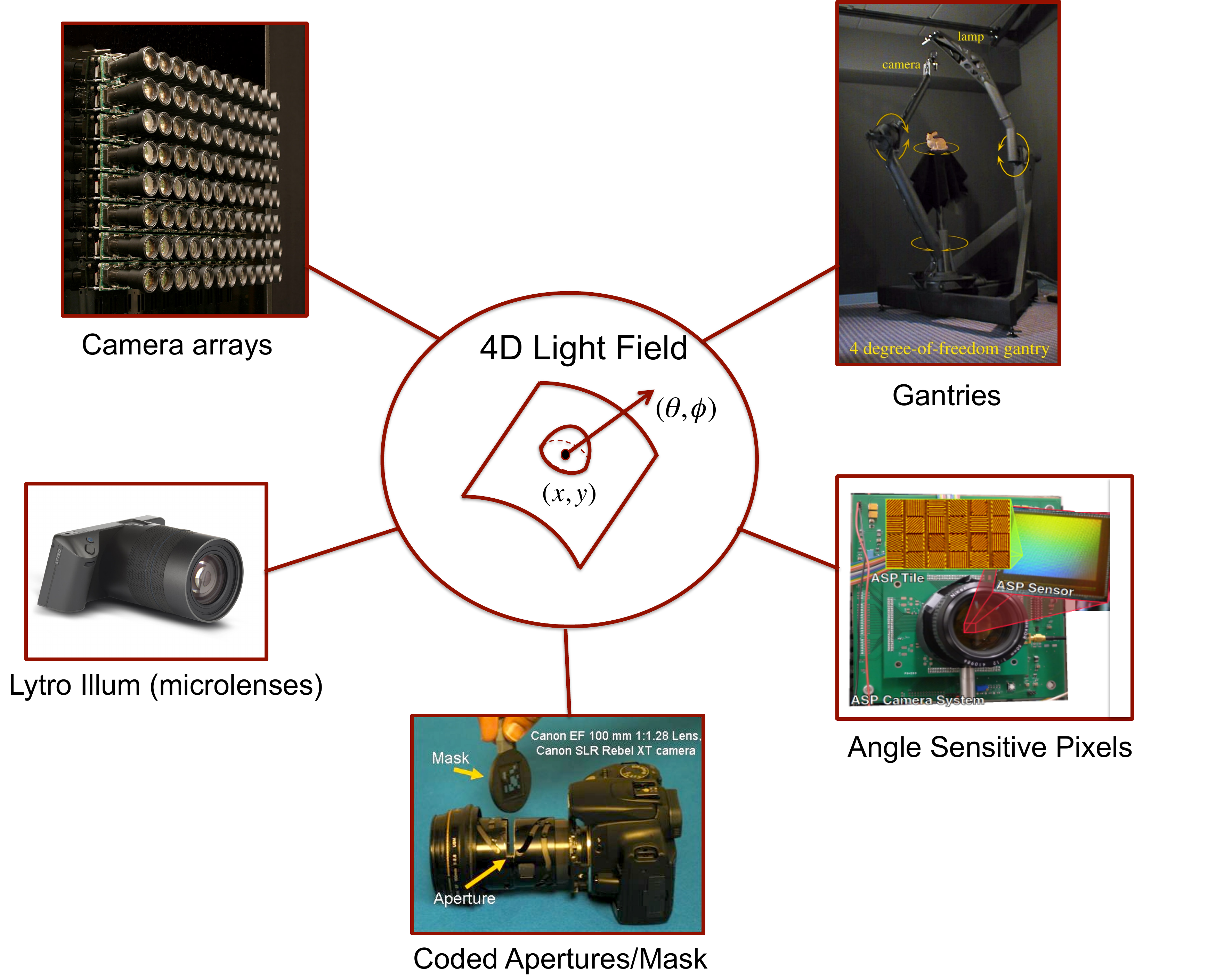}
\end{center}
   \caption{\textbf{Light Field Capture:} Light field capture has been performed with various types of imaging systems, but all suffer from challenges with sampling and processing this high dimensional information.}
\label{lightfieldconcept}
\end{figure}

In this section, we describe the image formation model for capturing 4D light fields and how to reconstruct them. 

A 4D light field is typically parameterised with either two planes or two angles~\cite{levoy1996light, gortler1996lumigraph}. We will represent light fields $l(x,y,\theta,\phi)$ with two spatial coordinates and two angular coordinates. For a regular image sensor, the angular coordinates for the light field are integrated over the main lens, thus yielding the following equation:
\begin{equation}
i(x,y) = \int_{\theta} \int_{\phi} l(x,y,\theta,\phi) d\phi d\theta,
\end{equation}
where $i(x,y)$ is the image and $l(x,y,\theta,\phi)$ is the light field. 

Single-shot light field cameras add a modulation function $\Phi(x,y,\theta,\phi)$ that weights the incoming rays~\cite{Wetzstein:PlenopticMultiplexing:2012}:

\begin{equation}
i(x,y) = \int_{\theta} \int_{\phi} \Phi(x,y,\theta,\phi) \cdot l(x,y,\theta,\phi) d\phi d\theta.
\end{equation}
When we vectorize this equation, we get $\vec{i} = \Phi\vec{l}$ where the $\vec{l}$ is the vectorized light field, $\vec{i}$ is the vectorized image, and $\Phi$ is the matrix discretizing the modulation function. Since light fields are 4D and images are 2D, this is inherently an underdetermined set of equations where $\Phi$ has more columns than rows. 

The matrix $\Phi$ represents the linear transform of the optical element placed in the camera body. This is a decimation matrix for lenslets, comprised of random rows for coded aperture masks, or Gabor wavelets for Angle Sensitive Pixels (ASPs).

\subsection{Reconstruction}

To invert the equation, we can use a pseudo-inverse $\vec{l} = \Phi^{\dagger}\vec{i}$, but this solution does not recover light fields adequately and is sensitive to noise~\cite{Wetzstein:PlenopticMultiplexing:2012}. Linear methods do exist to invert this equation, but sacrifice spatial resolution by stacking image pixels to gain enough measurements so that $\Phi$ is a square matrix. 

To recover the light field at the high spatial image resolution, compressive light field photography~\cite{marwah2013compressive} formulates the following $\ell_1$ minimization problem:
\begin{equation}
\min_{\alpha} || \vec{i} - \Phi D \alpha ||^{2}_{2} + \lambda ||\alpha ||_1
\end{equation}
where the light field can be recovered by performing $l = D\alpha.$ Typically the light fields were split into small patches of $9\times9\times5\times5$ $(x,y,\theta,\phi)$ or equivalently sized atoms to be processed by the optimization algorithm. Note that this formulation enforces a sparsity constraint on the number of columns used in dictionary $D$ for the reconstruction. The dictionary $D$ was learned using a set of million light field patches captured by a light field camera and trained using a K-SVD algorithm~\cite{aharon2006img}. To solve this optimization problem, solvers such as ADMM~\cite{boyd2011distributed} were employed. Reconstruction times ranged from several minutes for non-overlapping patch reconstructions to several hours for overlapping patch reconstructions.

\section{Deep Learning for Light Field Reconstruction}
We first discuss the datasets of light fields we use for simulating coded light field capture along with our training strategy before discussing our network architecture.

\subsection{Light Field Simulation and Training}

\begin{figure}
\begin{center}
\includegraphics[width=\linewidth]{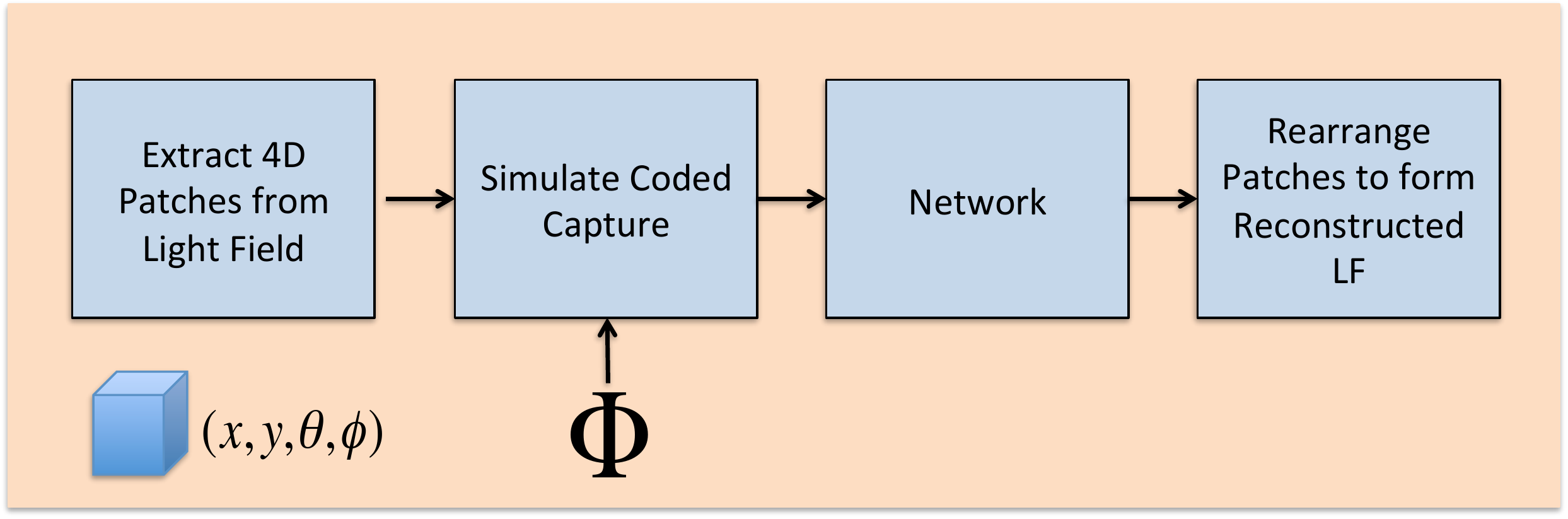} 
\end{center}
\caption{\textbf{Pipeline:} An overview of our pipeline for light field reconstruction.}
\label{Training}
\end{figure}

One of the main difficulties for using deep learning for light field reconstructions is the scarcity of available data for training, and the difficulty of getting ground truth, especially for compressive light field measurements. We employ a mixture of simulation and real data to overcome these challenges in our framework. 

\textbf{Synthetic Light Field Archive:}
We use synthetic light fields from the Synthetic Light Field Archive~\cite{Synthetic} which have resolution $(x,y,\theta,\phi) = (593,840,5,5)$. Since the number of parameters for our fully-connected layers would be prohibitively large with the full light field, we split the light fields into $(9,9,5,5)$ patches and reconstruct each local patch. We then stitch the light field back together using overlapping patches to minimize edge effects. This however does limit the ability of our network to use contextual light field information from outside this $(9,9,5,5)$ patch for reconstruction. However, as GPU memory improves with technology, we anticipate that larger patches can be used in the future with improved performance.

Our training procedure is outlined in Figure~\ref{Training}. We pick 50,000 random patches from four synthetic light fields, and simulate coded capture by multiplying by $\Phi$ to form images. We then train the network on these images with the labels being the true light field patches. Our training/validation split was 85:15. We finally test our network on a brand new light field never seen before, and report the PSNR as well as visually inspect the quality of the data. In particular, we want to recover parallax in the scenes, i.e. the depth-dependent shift in pixels away from the focal plane as the angular view changes. 

\textbf{Lytro Illum Light Field Dataset:} In addition to synthetic light fields, we utilize real light field captured from a Lytro Illum camera~\cite{LearningViewSynthesis}. To simulate coded capture, we use the same $\Phi$ models for each type of camera and forward model the image capture process, resulting in simulated images that resemble what the cameras would output if they captured that light field. There are a total of 100 light fields, each of size $(364,540,14,14)$. For our simulation purposes, we use only views $[6,10]$ in both $\theta$ and $\phi$, to generate $5x5$ angular viewpoints. We extract 500,000 patches from these light fields of size $(9,9,5,5)$, simulate coded capture, and use a training/validation split of 85:15. 

\subsection{Network Architecture}

\begin{figure*}
\includegraphics[height=6.5cm]{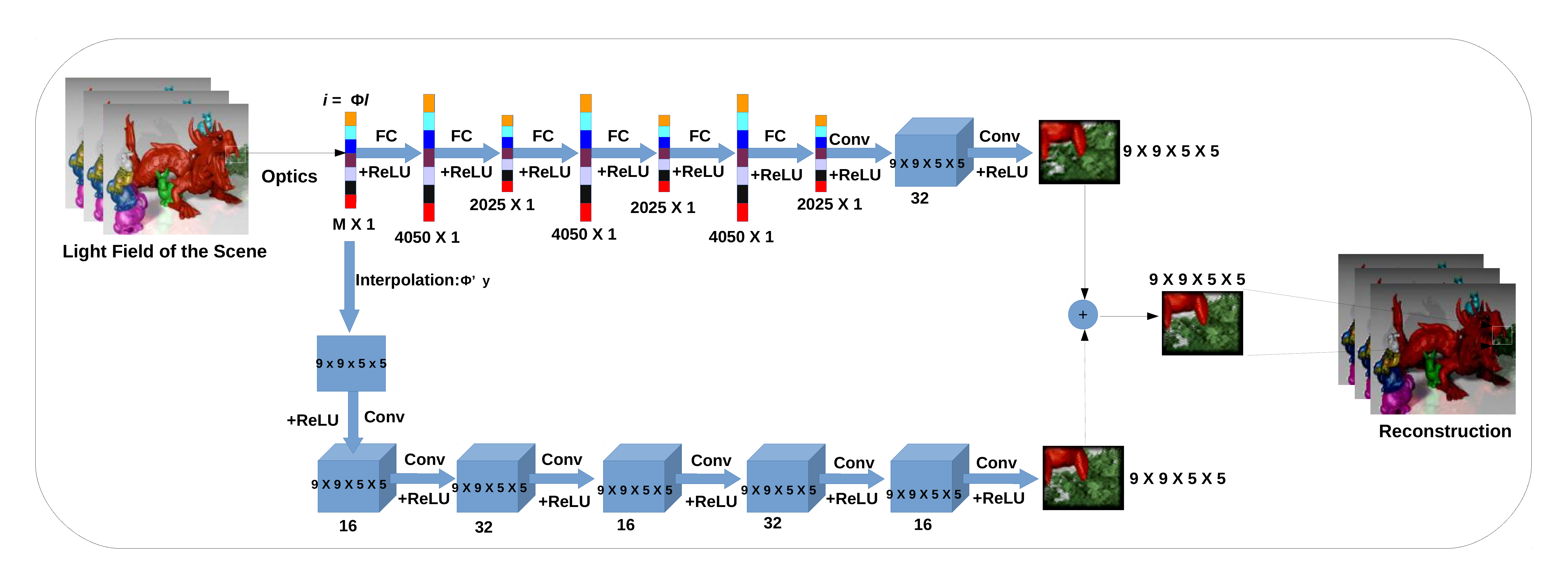}
\caption{\textbf{Network Architecture:} Our two branch architecture for light-field reconstruction. Measurements for every patch of size $(9,9,5,5)$ are fed into two parallel paths, one autoencoder consisting of 6 fully connected followed by one 4D convolution layer, and the other consisting of five 4D convolutional layers. The outputs of the two branches are added with equal weights to obtain the final reconstruction for the patch. Note that the size of filters in all convolution layers is $3\times3\times3\times3$.}
\label{fig:network}
\end{figure*}

Our network architecture consists of a two branch network, which one can see in Figure~\ref{fig:network}. In the upper branch,  the 2D input patch is vectorized to one dimension, then fed to a series of fully connected layers that form a stacked autoencoder (i.e. alternating contracting and expanding layers). This is followed by a 4D convolutional layer. The lower branch is a 4D CNN which uses a fixed interpolation step of multiplying the input image by $\Phi^T$ to recover a 4D spatio-angular volume, and then fed through a series of 4D convolutional layers with ReLU nonlinearities. Finally the outputs of the two branches are combined with weights of $0.5$ to estimate the light field. 

\begin{figure*}
\begin{center}
\includegraphics[height=6cm]{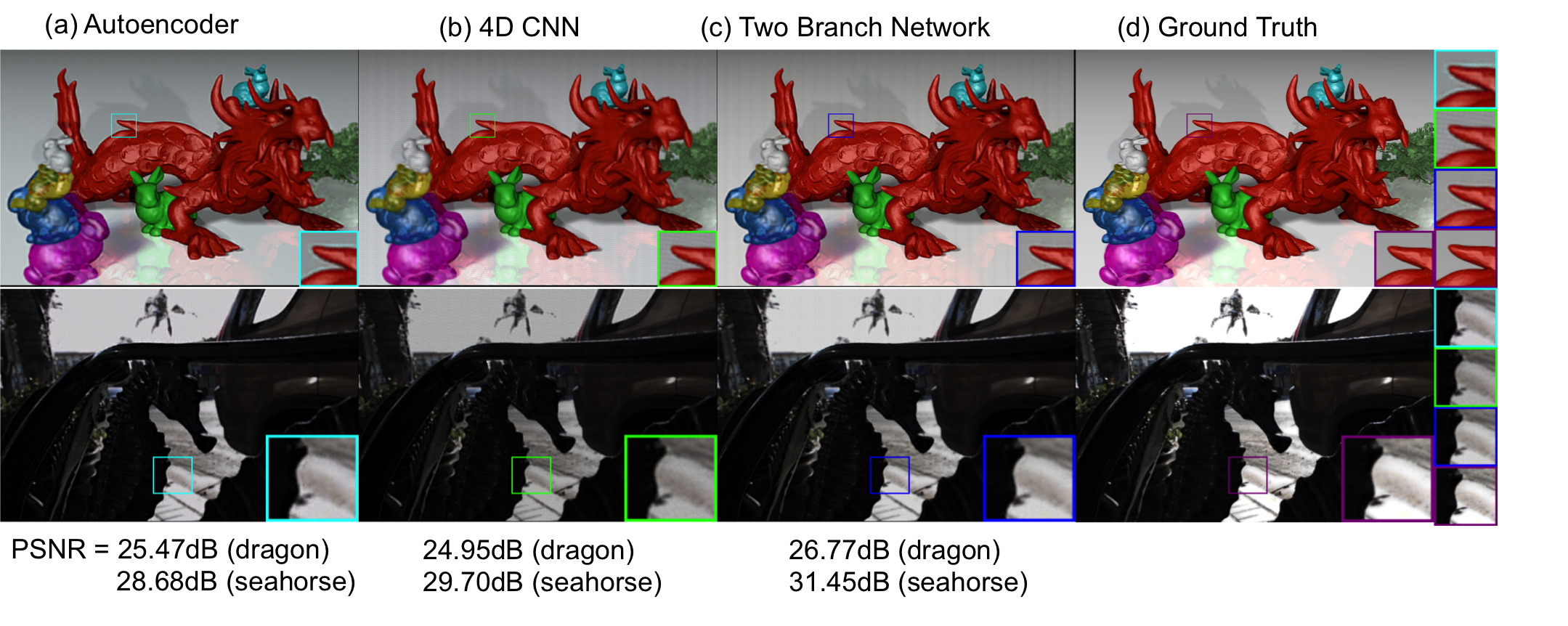}
\end{center}
\caption{\textbf{Branch Comparison:} We compare the results of using only the autoencoder or 4D CNN branch versus the full two branch network. We obtain better results in terms of PSNR for the two-stream network than the two individual branches. }
\label{fig:branchcompare}
\end{figure*}

There are several reasons why we converged on this particular network architecture.  Autoencoders are useful at extracting meaningful information by compressing inputs to hidden states~\cite{vincent2010stacked}, and our autoencoder branch helped to extract parallax (angular views) in the light field. In contrast, our 4D CNN branch utilizes information from the linear reconstruction by interpolating with $\Phi^T$ and then cleaning the result with a series of 4D convolutional layers for improved spatial resolution. Combining the two branches thus gave us good angular recovery along with high spatial resolution (please view the supplemental video to visualize the effect of the two branches). Our approach here was guided by a high-level empirical understanding of the behavior of these network streams, and thus, it is likely to be one of several architecture choices that could lead to similar results. In Figure~\ref{fig:branchcompare}, we show the results of using solely the upper or lower branch of the network versus our two stream architecture, which helped influence our design decisions. To combine the two branches, we chose to use simple averaging of the two branch outputs. While there may be more intelligent ways to combine these outputs, we found that this sufficed to give us a 1-2 dB PSNR improvement as compared to the autoencoder or 4D CNN alone, and one can observe the sharper visual detail in the inlets of the figure. 

\begin{figure}
\begin{center}
\includegraphics[width = \linewidth]{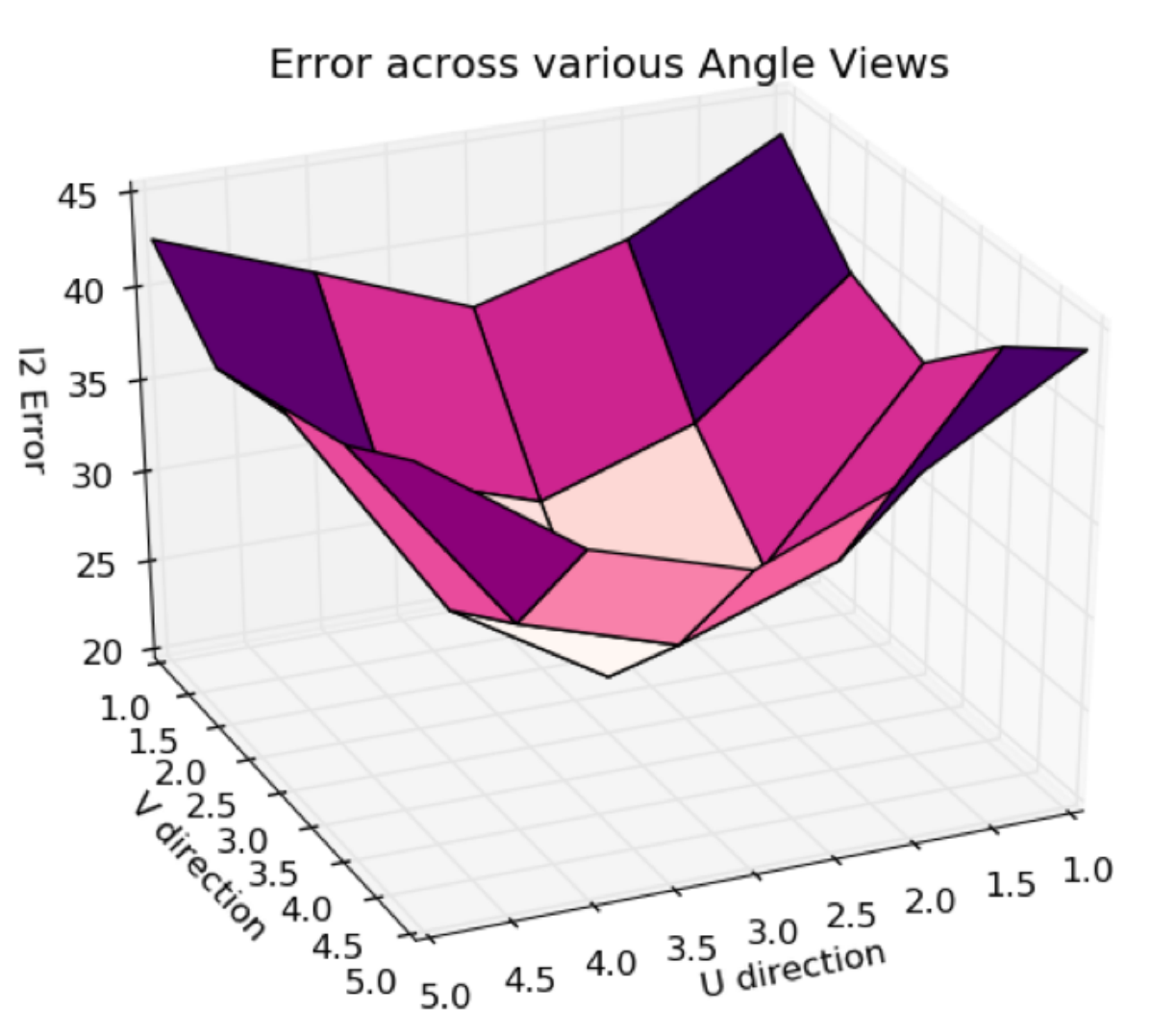}
\end{center}
\caption{\textbf{Error in Angular Viewpoints:} Here we visualize the $\ell_2$ error for a light field reconstruction with respect to ground truth using a standard $\ell_2$ loss function for training. Notice how the extreme angular viewpoints contain the highest error. This helped motivate the use of a weighted $\ell_2$ function for training the network. }
\label{fig:lossangle}
\end{figure}

For the loss function, we observed that the regular $\ell_2$ loss function gives decent reconstructions, but the amount of parallax and spatial quality recovered in the network at the extreme angular viewpoints were lacking. We note this effect in Figure~\ref{fig:lossangle}. To remedy this, we employ the following weighted $\ell_2$ loss function which penalizes errors at the extreme angular viewpoints of the light field more heavily:
\begin{equation}
L(l,\hat{l}) = \sum_{\theta,\phi} W(\theta,\phi)\cdot || l(x,y,\theta,\phi) - \hat{l}(x,y,\theta,\phi) ||_2^2,
\end{equation}
where $W(\theta,\phi)$ are weights that increase for higher values of $\theta,\phi$. The weight values were picked heuristically for large weights away from the center viewpoint with the following values: $W(\theta,\phi) = $ \[ \left( \begin{array}{ccccc}
 \sqrt{5}& 2 & \sqrt{3}& 2 & \sqrt{5} \\
2& \sqrt{3} & \sqrt{2} & \sqrt{3} & 2\\
\sqrt{3} & \sqrt{2} & 1 & \sqrt{2} & \sqrt{3} \\
2 & \sqrt{3} & \sqrt{2} & \sqrt{3} & 2 \\
\sqrt{5} & 2 & \sqrt{3} & 2 & \sqrt{5} \end{array} \right)\]  . This loss function gave an average improvement of $0.5$dB in PSNR as compared to $\ell_2$. 

\subsubsection{Training Details}

All of our networks were trained using Caffe~\cite{jia2014caffe} and using a NVIDIA Titan X GPU. Learning rates were set to $\lambda = .00001$, we used the ADAM solver~\cite{kingma2014adam}, and models were trained for about 60 epochs for 7 hours or so. We also finetuned models trained on different $\Phi$ matrices, so that switching the structure of a $\Phi$ matrix did not require training from scratch, but only an additional few hours of finetuning. 

For training, we found the best performance was achieved when we trained each branch separately on the data, and then combined the branches and jointly finetuned the model further on the data. Training from scratch the entire two branch network led to suboptimal performance of 2-3 dB in PSNR, most likely because of local minima in the loss function as opposed to training each branch separately and then finetuning the combination. 

\section{Experimental Results}
In this section, we show experimental results on both simulated light fields, real light fields with simulated capture, and finally real data taken from a prototype ASP camera~\cite{asp}. We compare both visual quality and reconstruction time for our reconstructions, and compare against baselines for each dataset. 

\subsection{Synthetic Experiments}
\begin{figure*}
\includegraphics[height=6cm]{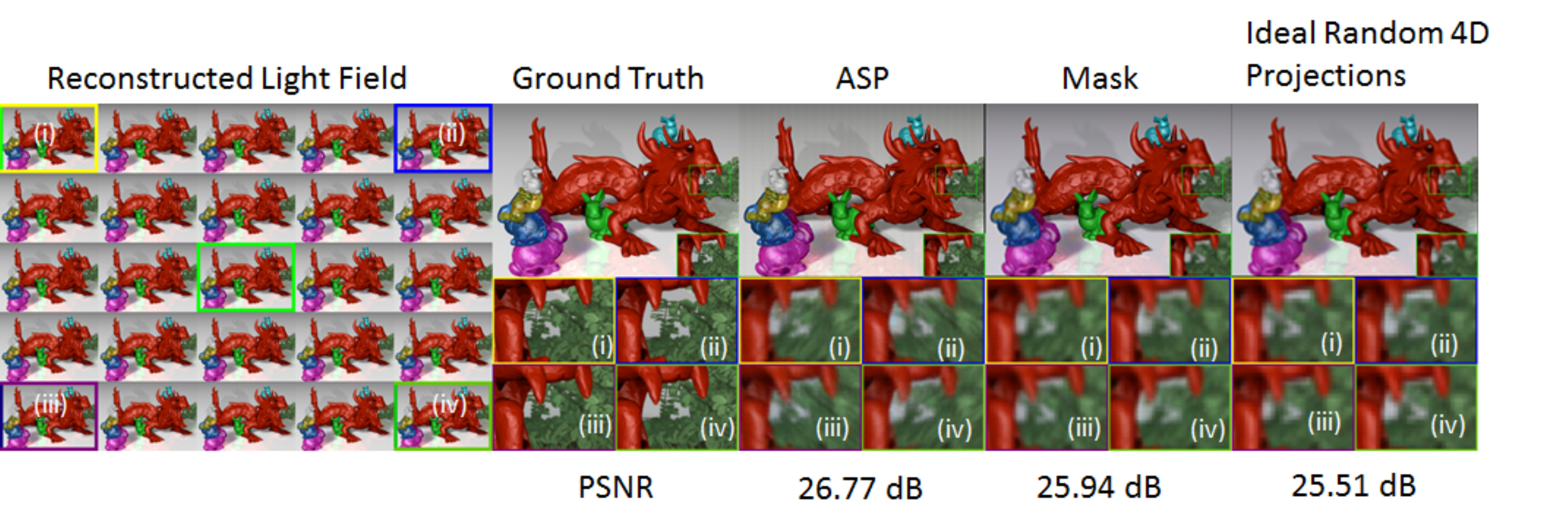}
\caption{\textbf{Different Camera Models:} We compare reconstructions for the dragons scene for different encoding schemes, ASP, Mask and Ideal Random 4D projections (CS) using the two branch network. These reconstructions were done at a low compression ratio of $8\%$ and with a stride of 5. At this low compression ratio, ASPs reconstruct slightly better (26.77 dB) as compared to Masks (25.96 dB) and CS (25.51 dB), although all methods are within ~1 dB of each other}
\label{fig:Phicompare}
\end{figure*}

We first show simulation results on the Synthetic Light Field Archive\footnote{Code available here: https://gitlab.com/deep-learn/light-field}. We used as our baseline the dictionary-based method from~\cite{marwah2013compressive, asp} with the dictionary trained on synthetic light fields, and we use the dragon scene as our test case. We utilize three types of $\Phi$ matrices, a random $\Phi$ matrix that represents the ideal 4D random projections matrix (satisfying RIP~\cite{candes2008restricted}), but is not physically realizable in hardware (rays are arbitrarily summed from different parts of the image sensor array). We also simulate $\Phi$ for coded masks placed in the body of the light field camera, a repeated binary random code that is periodically shifted in angle across the sensor array. Finally, we use the $\Phi$ matrix for ASPs which consists of 2D oriented sinusoidal responses to angle as described in~\cite{asp}. As can be seen in Figure~\ref{fig:Phicompare}, the ASPs and the mask reconstructions perform slightly better than the ideal random projections. It is hard to justify why ideal projections are not the best reconstruction in practice, but it might be because the compression ratio is too low at $8\%$ for random projections or because there are no theoretical guarantees that the network can solve the CS problem. All the reconstructions do suffer from blurred details in the zoomed inlets, which means that there is still spatial resolution that is not recovered by the network. 

\textbf{Compression ratio} is the ratio of independent coded light field measurements to angular samples to reconstruct in the light field for each pixel. This directly corresponds to the number of rows in the $\Phi$ matrix which correspond to one spatial location $(x,y)$. We show three separate compression ratios and measure the PSNR for ASP light field cameras in Table~\ref{fig:compressionsweep} with non-overlapping patches. Not surprisingly, increasing the number of measurements increased the PSNR. We also compared for ASPs using our baseline method based on dictionary learning. Our method achieves a 2-4 dB improvement over the baseline method as we vary the number of measurements.   
\begin{table}
\begin{center}
\resizebox{\columnwidth}{!}{%
\begin{tabular}{|c|c|c|}
\hline
\multicolumn{1}{|l|}{\textbf{Number of Measurements}} & \multicolumn{1}{l|}{\textbf{Our Method (PSNR)}} & \multicolumn{1}{l|}{\textbf{Dictionary Method (PSNR)}} \\ \hline
N = 2 & 25.40 dB & 22.86 dB \\ \hline
N = 15 & 26.54 dB & 24.40 dB \\ \hline
N = 25 & 27.55 dB & 24.80 dB \\ \hline
\end{tabular}%
}
\end{center}
\caption{\textbf{Compression sweep:} Variation of PSNR for reconstructions with the number of measurements in the dragons scene for ASP (non-overlapping patches) using the two branch network versus the dictionary method.}
\label{fig:compressionsweep}
\end{table}


\textbf{Noise:} We also tested the robustness of the networks to additive noise in the input images for ASP reconstruction. We simulated Gaussian noise of standard deviation of 0.1 and 0.2, and record the PSNR and reconstruction time which is display in Table~\ref{fig:noisesweep}. Note that the dictionary-based algorithm takes longer to process noisy patches due to its iterative $\ell_1$ solver, while our network has the same flat run time regardless of the noise level. This is a distinct advantage of neural network-based methods over the iterative solvers. The network also seems resilient to noise in general, as our PSNR remained about 26 dB.
\begin{table}
\begin{center}
    \begin{tabular}{||c|c|c|c|c||}
        \hline
        \textbf{Metrics} & \textbf{Noiseless} & \textbf{Std 0.1} & \textbf{Std 0.2}\\
        \hline \hline
        PSNR (Ours) [dB] & 26.77 & 26.74 & 26.66 \\
        \hline
        PSNR (Dictionary) [dB] & 25.80 & 21.98 & 17.40 \\

        \hline
        Time (Ours) [s] & 242 & 242 & 242 \\
        \hline
        Time (Dictionary) [s]& 3786 & 9540 & 20549 \\
        \hline
    \end{tabular}
\end{center}
\caption{\textbf{Noise:} The table shows how PSNR varies for different levels of additive Gaussian noise for ASP reconstructions. It is clear that our method is extremely robust to high levels of noise and provides high PSNR reconstructions, while for the dictionary method, the quality of the reconstructions degrade with noise. Also shown is the time taken to perform the reconstruction. For our method, the time taken is only 242 seconds and independent of noise level whereas for dictionary learning method, it can vary from 1 hour to nearly 7 hours.}
\label{fig:noisesweep}
\end{table}

\begin{figure}
\includegraphics[width=\columnwidth]{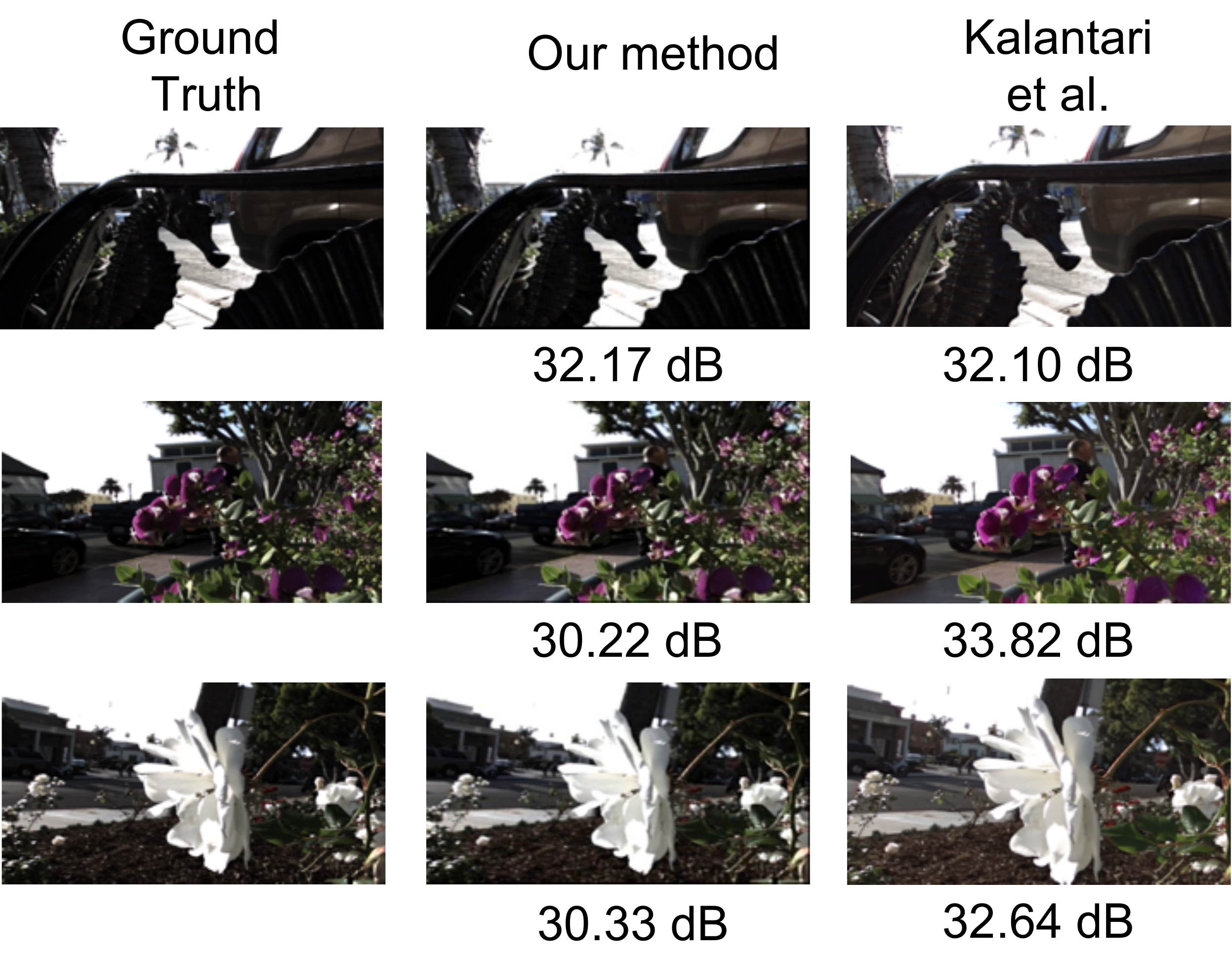}
\caption{\textbf{Lytro Illum Light Fields:} We show reconstruction results for real Lytro Illum light fields with simulated ASP capture. We note that our network performs subpar to Kalantari \etal~\cite{LearningViewSynthesis} since we have to deal with the additional difficulty of uncompressing the coded measurements. }
\label{fig:ucsd}
\end{figure}

\textbf{Lytro Illum Light Fields Dataset:} We show our results on this dataset in Figure~\ref{fig:ucsd}. As a baseline, we compare against the method from Kalantari \etal~\cite{LearningViewSynthesis} which utilize 4 input views from the light field and generate the missing angular viewpoints with a neural network. Our network model achieves higher PSNR values of 30-32 dB on these real light fields for ASP encoding while keeping the same compression ratio of $\frac{1}{16}$ as Kalantari \etal. While their method achieves PSNR $>$ 32dB on this dataset, their starting point is 4D light field captured by the Lytro camera and they do not have to uncompress coded measurements. In addition, our method is slightly faster as their network takes 147 seconds to reconstruct the full light field, while our method reconstructs a light field in 80 seconds (both on a Titan X GPU). 

%


\subsection{Real Experiments}

Finally, to show the feasibility of our method on a real compressive light field camera, we use data collected from a prototype ASP camera~\cite{asp}. This data was collected on an indoors scene, and utilized three color filters to capture color light fields. 

\begin{figure*}[!htbp]
\includegraphics[width=\textwidth]{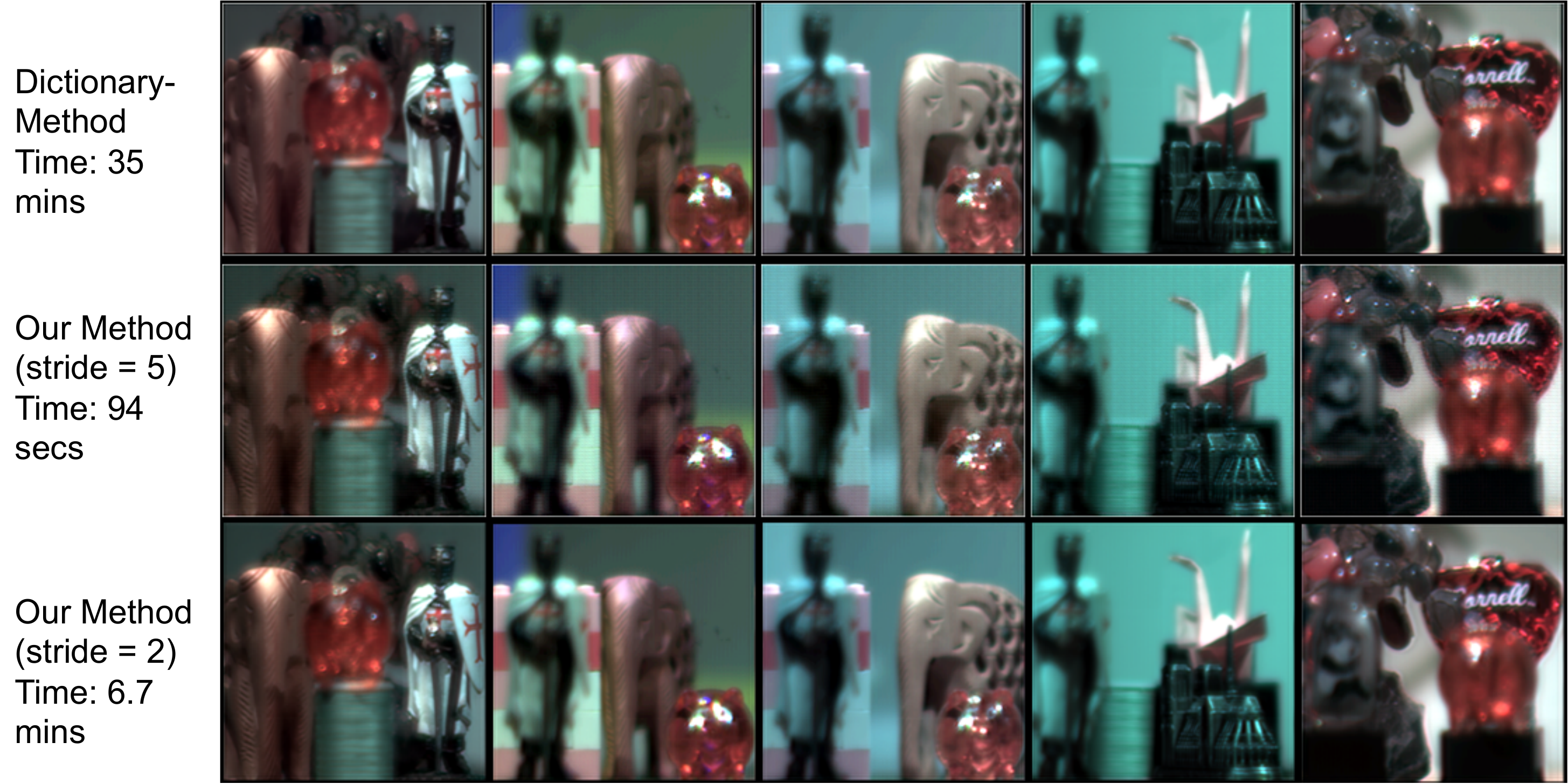}
\caption{\textbf{Real ASP Data:} We show the reconstructions for the real data from the ASP measurements using our method (for stride 5 and stride 2) and dictionary method (for stride 5), and the corresponding time taken. It is clear that the spatial resolution for our method is comparable as that using the dictionary learning method, and the time taken for our method (94 seconds) is an order less than that for the dictionary learning method (35 minutes). }
\label{fig:realexp}
\end{figure*}

Since we do not have training data for these scenes, we train our two branch network on synthetic data, and then apply a linear scaling factor to ensure the testing data has the same mean as the training data. We also change our $\Phi$ matrix to match the actual sensors response and measure the angular variation in our synthetic light fields to what we expect from the real light field. See Figure~\ref{fig:realexp} and the supplementary videos for our reconstructions. We compare our reconstructions against the method from Hirsch \etal~\cite{asp} which uses dictionary-based learning to reconstruct the light fields. For all reconstruction techniques, we apply post-processing filtering to the image to remove periodic artifacts due to the patch-based processing and non-uniformities in the ASP tile, as done in~\cite{asp}. 

\begin{figure}[!h]
\includegraphics[width=\columnwidth]{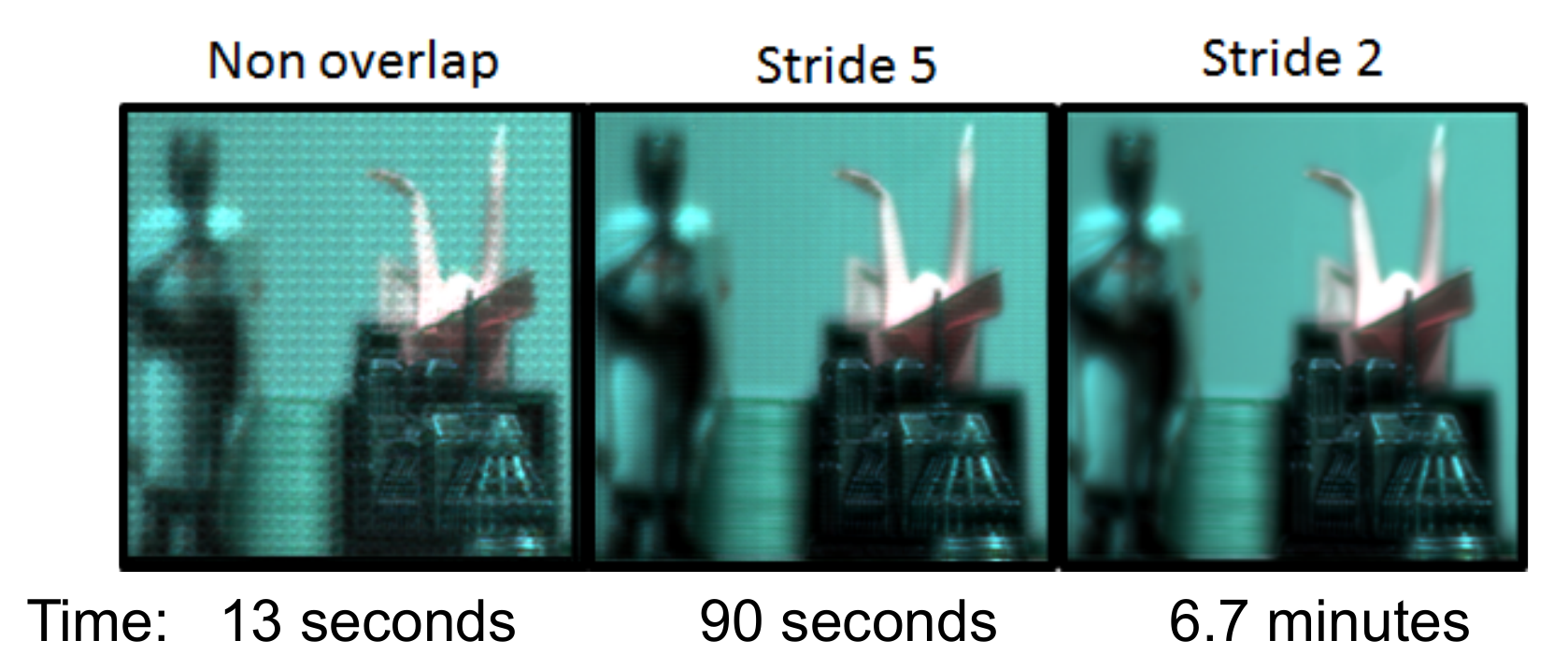}
\caption{\textbf{Overlapping Patches:} Comparison of non-overlapping patches and overlapping patches with strides of 11 (non-overlapping), 5, and 2 for light field reconstructions.}
\label{fig:overlap}
\end{figure}

We first show the effects of stride for overlapping patch reconstructions for the light fields, as shown in Figure~\ref{fig:overlap}. Our network model takes a longer time to process smaller stride, but improves the visual quality of the results. This is a useful tradeoff between visual quality of results and reconstruction time in general.

{\bf Time complexity and quality of ASP reconstructions:}
As can be seen, the visual quality of the reconstructed scenes from the network are on-par with the dictionary-based method, but with an order of magnitude faster reconstruction times. A full color light field with stride of 5 in overlapping patches can be reconstructed in 90 seconds, while an improved stride of 2 in overlapping patches yields higher quality reconstructions for 6.7 minutes of reconstruction time. The dictionary-based method in contrast takes 35 minutes for a stride of 5 to process these light fields. However, our method has some distortions in the recovered parallax that is seen in the supplementary videos. This could be possibly explained by several reasons. First, optical abberations and mismatch between the real optical impulse response of the system and our $\Phi$ model could cause artifacts in reconstruction. Secondly, the loss function used to train the network is the $l_2$ norm of the difference light field, which can lead to the well-known regress-to-mean effect for the parallax in the scene. It will be interesting to see if a $l_1$ based loss function or specially designed loss function can help improve the results. Thirdly, there is higher noise in the real data as compared to synthetic data. However, despite these parallax artifacts, we believe the results present here show the potential for using deep learning to recover 4D light fields from real coded light field cameras. 

\section{Discussion}
In this paper, we have presented a deep learning method for the recovery of compressive light fields that is signifcantly faster than the dictionary-based method, while delivering comparable visual quality. The two branch structure of a traditional autoencoder and a 4D CNN lead to superior performance, and we benchmark our results on both synthetic and real light fields, achieving good visual quality while reducing reconstruction time to minutes. 

\subsection{Limitations}

Since acquiring ground truth for coded light field cameras is difficult, there is no possibility of fine tuning our model for improved performance. In addition, it is hard to determine exactly the $\Phi$ matrix without careful optical calibration, and this response is dependent on the lens and aperture settings during capture time. All of this information is hard to feed into a neural network to adaptively learn, and leads to a mismatch between the statistics of training and testing data.

\subsection{Future Directions}
There are several future avenues for research. On the network architecture side, we can explore the use of generative adversarial networks~\cite{goodfellow2014generative} which have been shown to work well in image generation and synthesis problems~\cite{pathak2016context, ledig2016photo}. In addition, the network could jointly learn optimal codes for capturing light fields with the reconstruction technique, similar to the work by Chakrabarti~\cite{learningmultiplex} and Mousavi \etal~\cite{mousavi2015deep}, helping design new types of coded light field cameras. Finally, we could explore the recent unified network architecture presented by Chang \etal~\cite{chang2017one} that applies to all inverse problems of the form $y=Ax$. While our work has focused on processing single frames of light field video efficiently, we could explore performing coding jointly in the spatio-angular domain and temporal domain. This would help improve the compression ratio for these sensors, and potentially lead to light field video that is captured at interactive (1-15 FPS) frame rates. Finally, it would be interesting to perform inference on compressed light field measurements directly (similar to the work for inference on 2D compressed images~\cite{lohit2015reconstruction, kulkarni2015reconstruction}) that aims to extract meaningful semantic information. All of these future directions point to a convergence between compressive sensing, deep learning, and computational cameras for enhanced light field imaging.

\vspace{0.3cm}
\textbf{Acknowledgements:} The authors would like to thank the anonymous reviewers for their detailed feedback, Siva Sankalp for running some experiments, and Mark Buckler for GPU computing support. AJ was supported by a gift from Qualcomm. KK and PT were partially supported by NSF CAREER grant 1451263. SJ was supported by a NSF Graduate Research Fellowship and a Qualcomm Innovation Fellowship.


{\small
\bibliographystyle{ieee}
\bibliography{lfreconnet_review.bib}
}

\end{document}